\documentclass{article}

\usepackage{PRIMEarxiv}
\usepackage{todonotes}
\usepackage[utf8]{inputenc} %
\usepackage[T1]{fontenc}    %
\usepackage{hyperref}       %
\usepackage{url}            %
\usepackage{booktabs}       %
\usepackage{amsfonts}       %
\usepackage{nicefrac}       %
\usepackage{microtype}      %
\usepackage{lipsum}
\usepackage{fancyhdr}       %
\usepackage{graphicx}       %
\usepackage{CJKutf8}
\graphicspath{{media/}}     %

\usepackage{multirow}
  
\title{An Empirical Study of NetOps Capability of Pre-Trained Large Language Models
}

\newcommand\tab[1][0.7cm]{\hspace*{#1}}
\newcommand{\affa}{{$^{1}$}}
\newcommand{\affb}{{$^{2}$}}
\newcommand{\affc}{{$^{3}$}}
\newcommand{\affd}{{$^{4}$}}

\author{
Yukai Miao\affa \tab Yu Bai\affa \tab Li Chen\affa \tab Dan Li$^{1,2}$ \tab Haifeng Sun\affd \tab Xizheng Wang\affb \tab Ziqiu Luo\affb \\
\bf{Yanyu Ren}\affb \tab \bf{Dapeng Sun}\affa \tab \bf{Xiuting Xu}\affa \tab \bf{Qi Zhang}\affc \tab \bf{Chao Xiang}\affc \tab \bf{Xinchi Li}\affc \\
\affa Zhongguancun Laboratory \\
\affb Tsinghua University \\
\affc China Telecom Corporation Limited Research Institute \\
\affd Beijing University of Posts and Telecommunications
}

\begin{document}
\maketitle

\begin{abstract}

Nowadays, the versatile capabilities of Pre-trained Large Language Models (LLMs) have attracted much attention from the industry. However, some vertical domains are more interested in the in-domain capabilities of LLMs. For the Networks domain, we present NetEval\footnote{https://huggingface.co/datasets/NASP/neteval-exam}, an evaluation set for measuring the comprehensive capabilities of LLMs in Network Operations (NetOps). NetEval is designed for evaluating the commonsense knowledge and inference ability in NetOps in a multi-lingual context. NetEval consists of 5,732 questions about NetOps, covering five different sub-domains of NetOps. With NetEval, we systematically evaluate the NetOps capability of 26 publicly available LLMs. The results show that only GPT-4 can achieve a performance competitive to humans. However, some open models like LLaMA 2 demonstrate significant potential. \footnote{This is an on-going work, the contents are subject to change in the future.}

\end{abstract}

\section{Introduction}

\subsection{Empowering NetOps with LLMs}

Pre-trained Large Language Models (LLMs), are deep neural network models trained with massively large unlabeled textual data via self-supervised learning, obtaining comprehensive knowledge and generalization capabilities. With methods like fine-tuning and in-context learning, LLMs may serve as the foundation model for downstream tasks and achieve better performance and efficiency than traditional methods. In recent years, pre-trained LLMs have achieved remarkable progress in various tasks in general-domain Natural Language Processing tasks. Meanwhile, pre-trained LLMs are also applied in other domains like Software Engineering, Business and Finance, Biology and Medicine, etc. For NetOps, LLMs have enormous potential as well. The following are some of the potential application scenarios of LLMs in NetOps.

\begin{itemize}
\item Network Monitoring: Network monitoring software is deployed to monitor the performance data in networks in real-time, conduct analysis, generate reports, and quickly find and solve network faults to ensure the stability of the network.

\item Network Topology Planning: According to the needs of enterprises or organizations, network topology design and planning are carried out to ensure the rationality and scalability of the network structure. 

\item Network Device Management: Configure, install, monitor, maintain, upgrade, and test network equipment (such as switches, routers, firewalls, etc.) to ensure the normal operation of network equipment and avoid network failures and downtime.

\item Network Troubleshooting: Locate, analyze, and repair network faults to ensure rapid recovery and high availability.

\item Network Performance Optimization: Monitor and optimize performance indicators such as network bandwidth, latency, and packet loss rate to ensure stable and efficient operation of the network, and improve users' network experience and work efficiency.

\item Network Security: Carry out security assessment, vulnerability scanning, intrusion detection and defense on the network, protect the network from malicious attacks and data leakage, and ensure the security of the network.

\item Network Backup and Recovery: Back up network data regularly and make recovery plans to address unexpected data loss or catastrophic events, and safeguard essential data.
\end{itemize}

In summary, NetOps scenarios as above are crucial to the normal operation of enterprises and organizations. They not only ensure network stability and security but also enhance network performance and user experience, which makes greater value for enterprises and organizations. Leveraging the powerful representation and transfer capabilities, LLMs can effectively analyze and process NetOps data, thereby improve the efficiency and quality of NetOps and drive the high-quality development of intelligent NetOps. 

\subsection{Evaluation of NetOps Capabilities of LLMs}

Evaluating pre-trained LLMs for NetOps holds significant importance. On one hand, assessing LLMs' performance in NetOps provides a strong guidance for designing and optimizing LLMs for NetOps. On the other hand, such evaluation promote more applications of AI technology in NetOps, driving the further advancement of intelligent NetOps.

The evaluation of LLMs is by nature challenging. It is crucial to ensure the objectivity and fairness of the evaluation but often hard to achieve. The diverse nature of the tasks and data in NetOps presents additional challenges for evaluating pre-trained LLMs. To address these challenges, this work adopts the following methods to evaluate the capabilities of pre-trained LLMs in NetOps:

\begin{enumerate}
    \item Collect Data from Multiple Sources: As shown in Figure~\ref{fig:overview} , to address the comprehensiveness issue, multiple data sources that are related to NetOps are incorporated into the evaluation dataset. As the major component of the dataset, certification exam questions offer the advantage of not only thoroughly examining LLMs capabilities but also directly comparing LLMs with human performance, endorsing future application in real-world NetOps scenarios.

    \item Use Multi-format Questions as Evaluation Set: To address the objectiveness issue, we extract multi-choice questions from the data sources as the major part of evaluation set. Since multi-choice questions have deterministic answers, evaluating the correctness of the generated choice labels from LLMs can be highly automated. Besides, we also include some cloze questions and Closed-QA questions as a complement to the dataset.

    \item Diverse Prompt Engineering Methods to Enhance Model Output: Providing test questions directly as prompts to LLMs led to varied output quality, hindering most LLMs from performing at their full potential. To address this, we apply various prompt engineering techniques and rule-based post-processing to improve the evaluation results.
\end{enumerate}

This article is organized as follows: Section~\ref{sec:dataset} introduces the dataset we use for evaluation, Section~\ref{sec:models} elaborates the LLMs we evaluate, Section~\ref{sec:method} discusses the evaluation method, Section~\ref{sec:results} provides detailed evaluation results and analysis, and at last Section~\ref{sec:conclusion} concludes the paper.

\section{Evaluation Dataset}\label{sec:dataset}

\begin{figure}
    \centering
    \includegraphics[width=\linewidth]{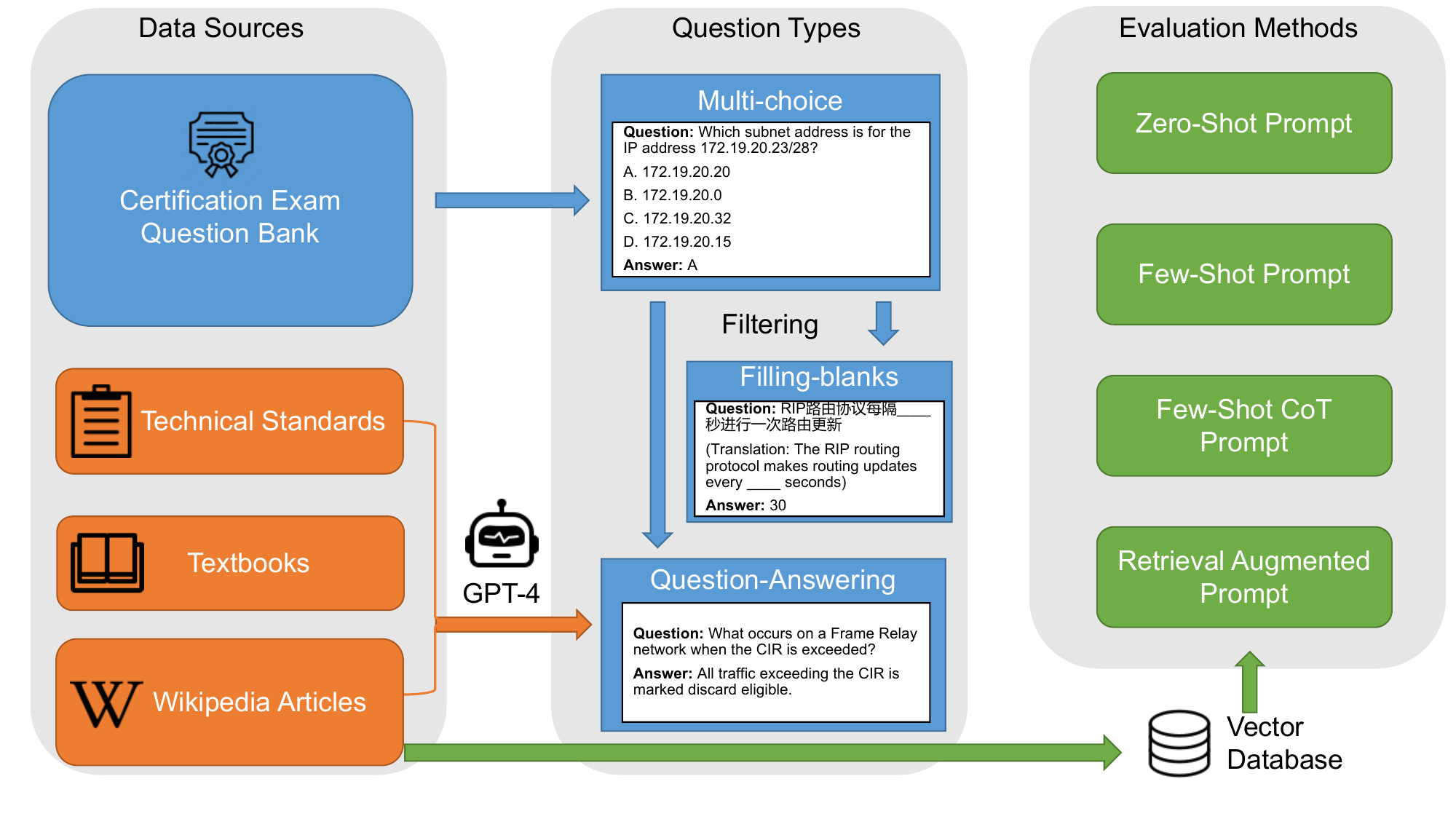}
    \caption{Overview of Evaluation Data Collection and the Evaluation Methods}
    \label{fig:overview}
\end{figure}

We present a multi-lingual LLM evaluation dataset for NetOps, named ``\textbf{NetEval}''. It consists of more than six thousand questions about NetOps, most of which are multi-choice questions. The construction process of NetEval is illustrated in Figure~\ref{fig:overview}. NetEval has been released to the public and will be continuously updated.

\subsection{Multi-choice Evaluation Set}

Multiple-choice questions are the most widely used questions in both real-world certification exams and existing LLM benchmarks. Many existing LLM evaluation datasets, e.g. the Massive Multitask Language Understanding (MMLU) dataset~\cite{hendrycks2020measuring}, are in the form of multiple-choice questions. Most of the recent proposed LLMs, such as DeepMind's Gopher and Chinchilla models, include MMLU in their evaluations.

\begin{figure}[h]
  \centering
  \includegraphics[width=0.5\textwidth]{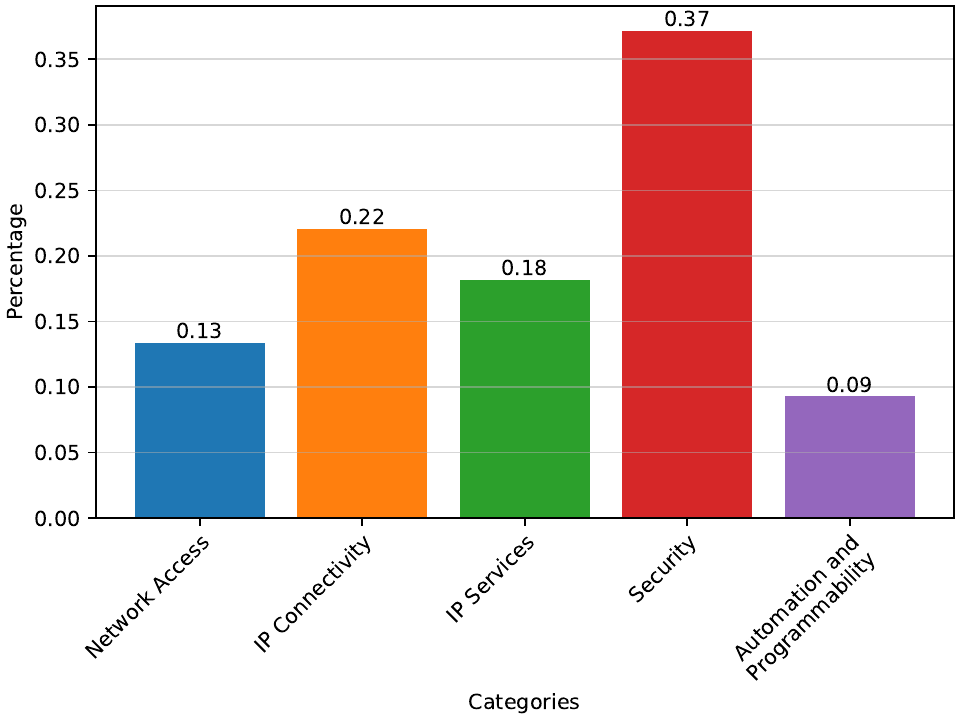} %
  \caption{Categories and the Question Distribution of Multi-choice Evaluation Set}
  \label{fig:categories}
\end{figure}

Inspired by MMLU, we construct a comprehensive multi-choice problem set for the NetOps domain by collecting certification exam questions from both the public websites and network device vendors we collaborate with. We argue that although answering multi-choice questions is far from solving real NetOps problems, achieving a human-level accuracy in NetOps certification exam questions is still one of the fundamental capabilities we expect LLMs to have. 
The collected evaluation set includes 5,269 multiple-choice questions. As show in in Figure~\ref{fig:categories}, we sample some of the questions and annotate the category each question is within from 5 distinct subjects. Table~\ref{tab:question-examples} shows the example question in each subject. Due to the varied data sources, the evaluation set contains both English questions (73\%) and Chinese questions (27\%), which facilitates  the evaluation of multi-lingual NetOps capability.

\begin{table}[]
\resizebox{\columnwidth}{!}{
\begin{tabular}{@{}ccc@{}}
\toprule
\textbf{Subject}           & \textbf{Question}               & \textbf{Choices} \\ 
\midrule
Network Access & Which MTU size can cause a baby giant error?
	& A. 1500 \, B. 9216 \, C. 1600 \, D. 1518 \\
\midrule
\multirow{4}{*}{IP Connectivity} & & A. IP routing must be enabled to allow the two hosts to communicate.\\
& Two hosts are attached to a switch with the default configuration. & B. The two hosts are in the same broadcast domain.\\
& Which statement about the configuration is true? & C. The switch must be configured with a VLAN to allow the two hosts to communicate.\\
& & D. Port security prevents the hosts from connecting to the switch.\\
\midrule
IP Services & Which NAT term is defined as a group of addresses available for NAT use? & A. NAT pool \, B. dynamic NAT \, C. static NAT \, D. one-way NAT\\
\midrule
\multirow{3}{*}{Security} & \begin{CJK}{UTF8}{gbsn}以下哪个攻击可以提供拦截和修改HTTP数据包功能？\end{CJK} & \multirow{3}{*}{A. Metasploit \, B. Hackbar \, C. Sqlmap \, D. Burpsuite}\\
& (Translation: Which of the following attacks is able & \\
& to intercept and modify HTTP packets?) & \\
\midrule
& An administrator is in the process of changing the configuration of a router. & A. Router\# show startup-config\\
Automation and & What command will allow the administrator to check the changes that & B. Router\# show current-config\\
Programmability &  have been made prior to saving the new configuration? & C. Router\# show running-config\\
& & D. Router\# show memory\\

\bottomrule
\end{tabular}
}
\vspace{.1cm}
    \caption{Example question in each subject}
    \label{tab:question-examples}
\end{table}

Regarding the organization of the dataset, the questions in each subject are divided into three parts: the development set, the validation set, and the test set. The development set includes five examples for 5-shot evaluation, as will be discussed in Section~\ref{sec:method}. The validation set is used for hyperparameter tuning, while the test set is used for the final evaluation.

To avoid the development set and the validation set having similar questions to that in test set, we select questions that were least similar to other questions as the development set and validation set. To measure the similarity between questions, we use sentence-BERT~\cite{reimers2019sentence} to encode each question sample into a vector and calculate the cosine similarity between vectors as the measurement of semantic similarity between samples. 

We release all the development sets and validation sets of the multi-choice questions, while the test sets are only partially available to the public. The users of this dataset need to submit the request of evaluation to obtain the results on the full test set.

\subsection{Non-multi-choice Evaluation Set}

The evaluation solely on the multi-choice questions has some obvious limitations. 
First, the LLMs may guess the answer labels, regardless how much NetOps knowledge does it have. 
Second, the lack of interpretability of multi-choice question-answering hinders finer-grained evaluation and in-depth analysis. 
Thus, as a complement to our evaluation set, we also include a small collection of non-multi-choice questions, consisting of 463 filling-blanks and question-answering questions. 

We adopt two ways to generate these questions. First, we use rule-based filtering method to extract a few questions from the multi-choice question set, and eliminate the choices to obtain a filling-blanks or question-answering questions. The filtering rules are manually crafted to identify the questions which are still valid given only the question content and the correct choice. Second, we also adopt GPT-4 to automatically extract question-answering pairs from the raw corpus about NetOps, including the technical standards (e.g. RFC), networking textbooks and Wikipedia articles that are related to NetOps. We parse the raw corpus and split them into small chunks (at most 4k characters in each chunk), combine each chunk with a piece of instruction about the requirement of the extraction output, and finally send the instructions to GPT-4 via the API. We also adopt similar methods in Alpaca to post-process the response from GPT-4 to extract the expected output. To shortlist the Wikipedia articles that are relevant to NetOps, we manually create a list of seed Wikipedia pages of NetOps terms, e.g. IPv6, and then expand the article list by traversing the hyperlinks in the articles in the current list. In this way, we collected around two thousand articles.

After the collection of non-multi-choice questions, we further filter the questions with manual inspection and eventually select 463 questions to do the evaluation. However, since the inference on such questions requires a larger generation length, resulting in a much larger cost, we only evaluate some selected models in Table~\ref{tab:llms} on this evaluation set.

We organize the non-multi-choice questions in a similar way as that for multi-choice questions, i.e. split the questions into the development set, validation set and the test test. Due to the potentially fast change in he non-multi-choice questions. We choose not to release them for now, but we are happy to share them upon request.

\section{Evaluated Models}\label{sec:models}

With the rapid advances of LLM technology and the vast amount of investment in LLM from various institutions and companies, new LLMs keep emerging in the market. For example, at the time of writing this article, there have been more than 100 LLMs developed in China. We are not able to cover all the models in our evaluation due to the limited computational resources. Instead, we select the most representative ones in our experiments.

As shown in Table~\ref{tab:llms}, we evaluate general-domain models from 6 institutions or companies. OpenAI developed many huge LLMs with hundreds or thousands of billions of parameters. Among these models, text-davinci-003 is the 175B base model in GPT-3.5 series, GPT-3.5-turbo~\cite{openai2023} is a chat-oriented model obtained by supervised finetuning (SFT) and reinforcement learning from human feedback (RLHF) on top of text-davinci-003, and the successor GPT-4~\cite{openai2023gpt} is recognized as the most advanced LLM that ever exists. OpenAI does not provide the model weights to the public but exposes some APIs to the developers, thus we evaluate these models via the APIs. As a good open alternative to GPT-3/4 series, LLaMA~\cite{touvron2023llama1} was introduced by Meta, and attracted much attention from both the academia and the industry. There are multiple sizes available in LLaMA series, ranging from 7B to 65B. Its successor LLaMA 2~\cite{touvron2023llama2} provides both the base models and chat-oriented models of 3 different sizes. Falcon~\cite{penedo2023refinedweb} was developed by UAE TII, consisting of both the base models and the instruction-tuned models of two sizes, once ranked the first in the Open LLM Leaderboard~\cite{open-llm-leaderboard}. GLM~\cite{du2021glm} is a series of base models from Tsinghua University, trained with a distinct architecture and bilingual (English+Chinese) corpus. The chat-variants of GLM, including ChatGLM and ChatGLM 2, are also derived from the base model with SFT and RLHF. Besides, the Moss~\cite{fudan2023moss} and Baichuan~\cite{baichuan2023} are also competitive LLMs that pre-trained with bilingual corpus.

\begin{table}[]
\resizebox{\columnwidth}{!}{
\begin{tabular}{@{}cccccc@{}}
\toprule
\textbf{Organization}           & \textbf{Series}               & \textbf{Name}                  & \textbf{Model Size}   & \textbf{Model Type} & \textbf{Open Weights} \\ \midrule
\multirow{3}{*}{OpenAI}  & \multirow{3}{*}{GPT}      & text-davinci-003~\cite{openai2023}        & 175B           & Base          & No             \\
                         &                           & GPT-3.5-turbo~\cite{openai2023}        & 175B           & Dialog          & No             \\
                         &                           & GPT-4~\cite{openai2023gpt}                & Unknown    & Dialog          & No             \\
\midrule
\multirow{3}{*}{Meta}    & \multirow{3}{*}{LLaMA}    & LLaMA~\cite{touvron2023llama1}                & 7B/13B/30B/65B & Base          & Yes (Need Authorization)        \\
                         &                           & LLaMA 2~\cite{touvron2023llama2}              & 7B/13B/70B     & Base          & Yes (Need Authorization)        \\
                         &                           & LLaMA 2 Chat~\cite{touvron2023llama2}         & 7B/13B/70B     & Base          & Yes (Need Authorization)        \\
\midrule
\multirow{2}{*}{UAE TII} & \multirow{2}{*}{Falcon}   & Falcon~\cite{penedo2023refinedweb}               & 7B/40B         & Base          & Yes             \\
                         &                           & Falcon-instruct~\cite{penedo2023refinedweb}      & 7B/40B         & Dialog          & Yes             \\
\midrule
\multirow{3}{*}{Tsinghua University}    & GLM                       & GLM~\cite{du2021glm}                  & 10B/130B       & Base          & Yes             \\
                         & \multirow{2}{*}{ChatGLM}  & ChatGLM~\cite{du2021glm}              & 6B             & Dialog          & Yes             \\
                         &                           & ChatGLM 2~\cite{du2021glm}            & 6B             & Dialog          & Yes             \\
\midrule
\multirow{2}{*}{Fudan University}    & \multirow{2}{*}{Moss}     & Moss~\cite{fudan2023moss} (moon-003-base) & 16B            & Base          & Yes             \\
                         &                           & Moss~\cite{fudan2023moss} (moon-003-sft)   & 16B            & Dialog          & Yes             \\
\midrule
\multirow{2}{*}{Baichuan Inc.}    & \multirow{2}{*}{Baichuan} & Baichuan~\cite{baichuan2023}            & 7B/13B         & Base          & Yes             \\
                         &                           & Baichuan-Chat~\cite{baichuan2023}       & 13B            & Dialog          & Yes             \\ 
\bottomrule
\end{tabular}
}
\vspace{.1cm}
\caption{Evaluated Pre-trained LLMs}
\label{tab:llms}
\end{table}

\section{Evaluation Method}\label{sec:method}

\subsection{Evaluation Pipeline}

The evaluation process involves three steps, i.e. prompt generation, model inference, and post-processing. The main difference is in the prompt generation.

\begin{figure}
    \centering
    \includegraphics[width=\textwidth]{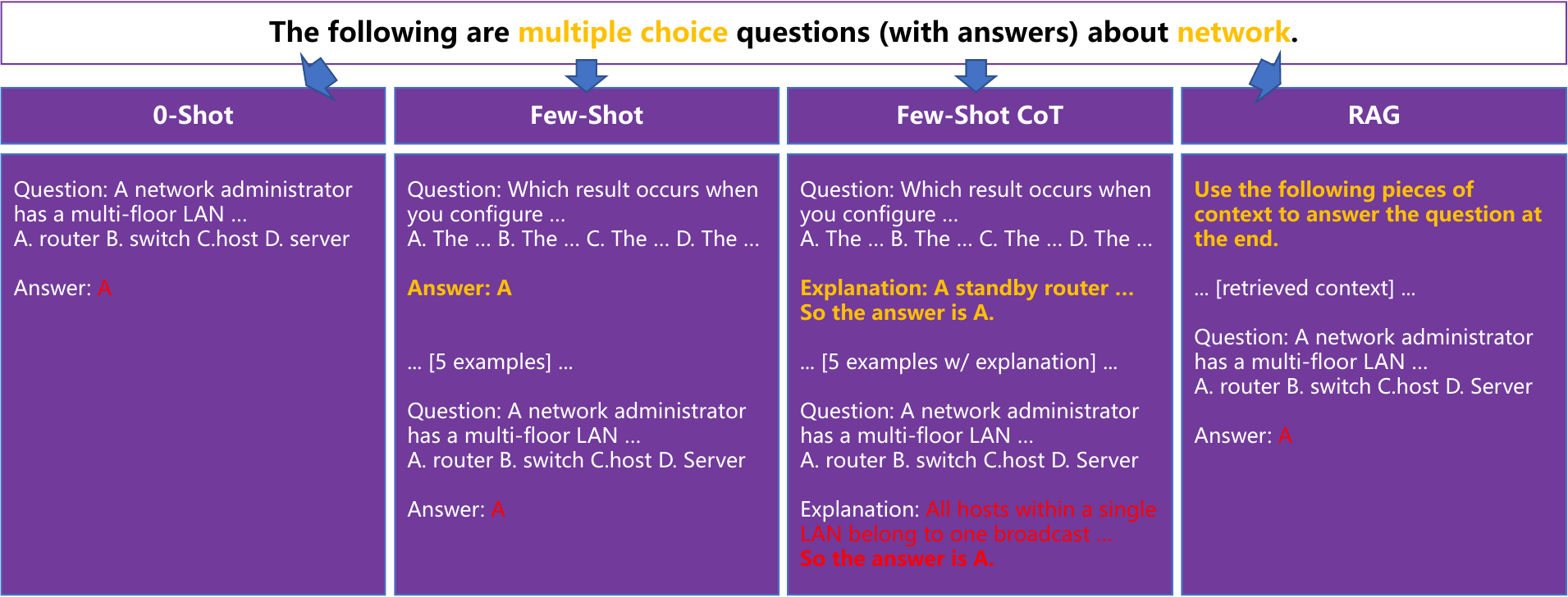}
    \caption{Different ways of Construction of Prompts}
    \label{fig:enter-label}
\end{figure}

\textbf{Prompt Generation}: In-context Learning (ICL) allows LLMs to perform tasks by observing a few prompt examples, without requiring any updates to the model parameters. Prompt generation involves transforming the questions from the evaluation set into textual input for LLM inference. The generated prompts mainly contain task descriptions, the content of the evaluation, and the final question. 
The task description presents the context of task, for example, ``The following are single-choice questions related to NetOps...''. As illustrated in Figure~\ref{fig:enter-label}, we adopt four types of prompts, namely zero-shot prompts, few-shot prompts, few-shot Chain-of-Thought (CoT) prompt and Retrieval-Augmented Generation (RAG) prompt. 

Zeor-shot and Few-shot prompts are two commonly used prompts in various LLM tasks. Zero-shot prompts contain only the content of the task description and test question, while few-shot prompts also include some examples with questions and answers. We use the development set in the dataset for sampling few-shot examples.

Most studies indicate that Chain of Thought (CoT) \cite{wei2022chain} has a significant impact on the output results of generative models. Therefore, we also consider the few-Shot Chain-of-Thought (Few-shot-CoT) prompts in our evaluation. Few-shot-CoT provides a small number of examples of chain of thought demonstrations to guide the reasoning process of the language model and consequently affect the inference results. We do not use zero-shot CoT as we observe extremely bad performance on all the evaluated models in our experiments.

Moreover, we also include Retrieval-Augmented Generation (RAG) prompt in our evaluation. In the field of prompt engineering, leveraging information retrieval techniques such as vector retrieval has become a prominent approach to enhancing the memory of LLMs. Semantic indexing allows for the swift and accurate retrieval of a set of text pieces from an extensive pool of candidate data that are semantically relevant to a specific task. In scenarios involving the addition of new samples, model retraining is unnecessary. Instead, by providing retrieval-based factual information to the model through prompts, its generation capability can be augmented. In our experiments, considering the computational cost, we only evaluate zero-shot RAG prompt.

At the end of the prompts, we used text like "Answer:" to indicate that the model should output the answer afterward.

\textbf{Model Inference}: After generating prompts, we invoked the interfaces of various language models to obtain model outputs.
For non-open source language models, such as ChatGPT, we achieved model inference by calling the API. Each evaluation question was processed individually as an individual session. The API input conversation only comprising a default system message and a user message containing the prompt content. For open source models, we load and test the model in the local environment. Depending on the implementation of the models, we call appropriate functions for text generation to conduct the inference of the model. When we call such functions, we refer to the example parameter configurations in the official sites of the models and restrict the maximal number of output tokens.

\textbf{Post-Processing}: Since all the evaluated LLMs are generative models, it is hard to unify the output formats. Post-processing is required to extract final answers from the model outputs. Initially, we truncate the output content using the newline character and retain only the first line. For multi-choice questions, it is often insufficient to extract the first line of the output. For example, the first line of the output might be 'the correct choice is A'. In cases as such, a series of predefined regular expressions are employed to match and extract the answer label from the generated text.

Upon completing the above steps, we compute the performance metrics on the model output. For multi-choice questions, we compute the accuracy of answers by directly comparing the extracted answer label with the ground truth. For non-multi-choice questions, we adopt BLEU\cite{Papineni2002BleuAM} and ROUGE\cite{Lin2004ROUGEAP} scores to evaluate the quality of the output.

\subsection{Implementation Details}

Our evaluation method is fundamentally similar to existing mainstream methods such as MMLU and C-Eval. 
For open source models, we loaded and tested the models in a local test environment. Depending on the implementation of the evaluated models, we call the low-level text generation functions or high-level pipeline functions to execute the inference given the prompt. %
When calling local models, we referred to parameter combinations in example inference code for each model in their official websites and constrained the number of output tokens. Our evaluation used two types of hosts as the testing environment as depicted in Table~\ref{tab:env}. For larger models such as LLaMA-65B and Falcon 40B, we used the Type B hosts for testing, while other models were dynamically allocated to all the hosts according to resource availability. The model inference was accelerated with model parallelism, which distribute the computation into multiple GPUs.

\begin{table}[]
\centering
\begin{tabular}{@{}ccccc@{}}
\toprule
\textbf{Type} & \textbf{Nodes} & \textbf{GPU}                   & \textbf{CPU}                                       & \textbf{Main Memory}    \\
\midrule
A    & 2    & NVIDIA A6000 48GB * 4 & Intel(R) Xeon(R) Gold 5218R CPU @ 2.10GHz & 256GB \\
B    & 1    & NVIDIA A100 80GB * 8  & AMD EPYC 7H12 64-Core Processor @ 2.6GHz  & 1TB  \\
\bottomrule
\end{tabular}
\vspace{.1cm}
\caption{Environment for Model Evaluation}
\label{tab:env}
\end{table}

The Retrieval-Augmented Generation relies on an external vector database and an efficient vector engine. In our evaluation, we adopt faiss~\cite{johnson2019billion} as the basis of the vector engine, combined with customized sharding strategy. We use a pre-built Wikipedia index, built with the sentence-bert~\cite{reimers2019sentence} encoder, as the vector database.

\section{Results and Analysis}\label{sec:results}

Since most of the questions in our evaluation set are multi-choice questions, we put most of our efforts on evaluating the LLMs on multi-choice questions only. For non-multi-choice questions, we examine selected models with zero-shot prompt only. We leave further experiments and full-scale evaluations on these questions as future work.

\begin{figure}[h]
  \centering
  \includegraphics[width=0.8\textwidth]{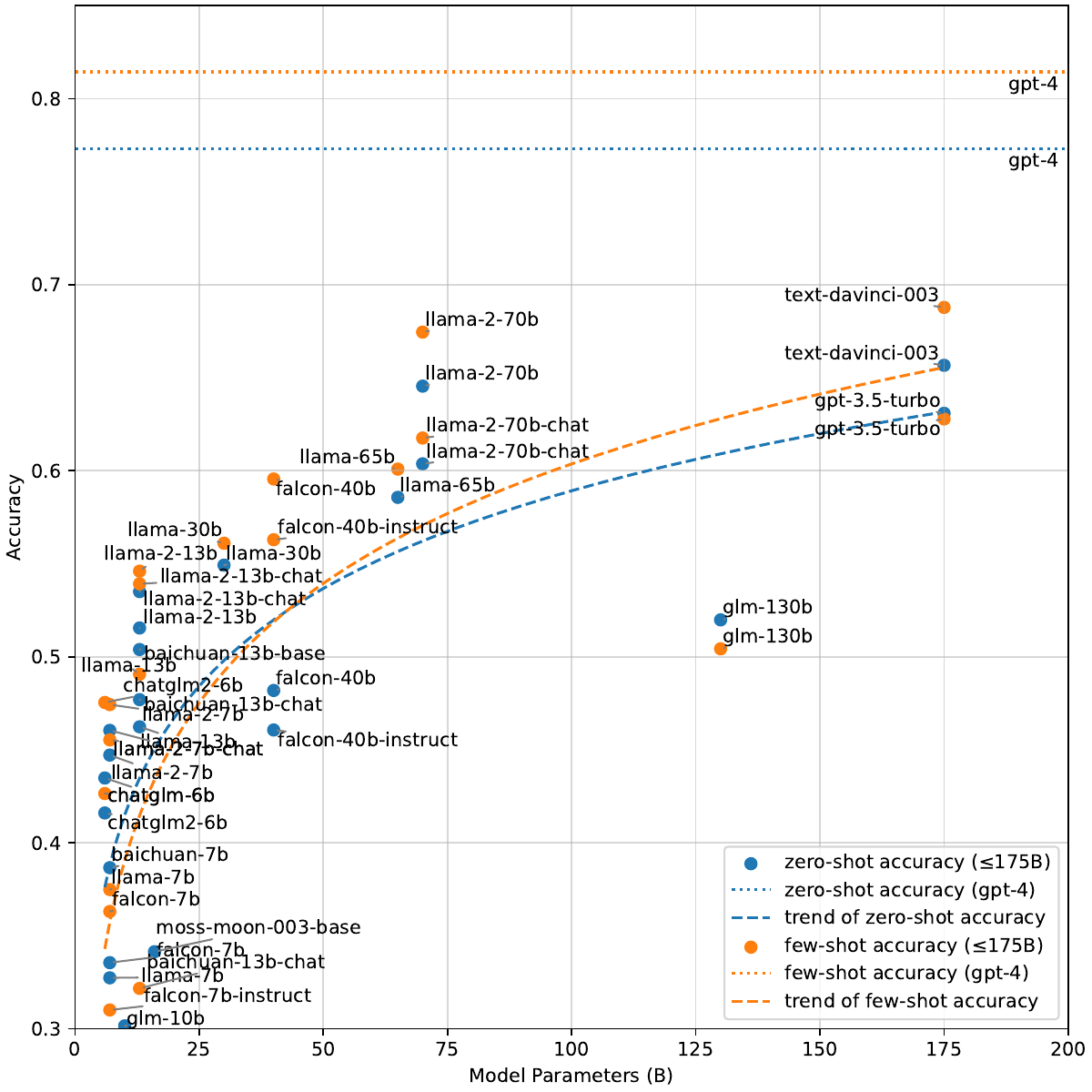} %
  \caption{Model Accuracy (Models with accuracy below 30\% are not displayed in this figure)}
  \label{fig:image2}
\end{figure}

The accuracy of various models under zero-shot and few-shot prompts is illustrated in Figure \ref{fig:image2}. Overall, GPT-4 achieves the best performance, i.e. 81\% accuracy under few-shot prompts and 77\% under zero-shot prompts. NetOps related certification exams require an 80\% passing threshold. In our test, GPT-4 (few-shot) is the only model that can pass the certification. The other models exhibited lower accuracy levels, preventing them from passing the human-certified exam.
We consider the main reason is that other models lack pre-training and fine-tuning on specialized NetOps corpora. However, open models like LLaMA 2 showcased considerable potential. %
Consequently, we deem it essential to further train LLMs specialized in the NetOps domain based on general open-source LLMs.

\subsection{Relationship between Model Parameters and Accuracy}
\begin{figure}[h]
  \centering
  \includegraphics[width=0.7\textwidth]{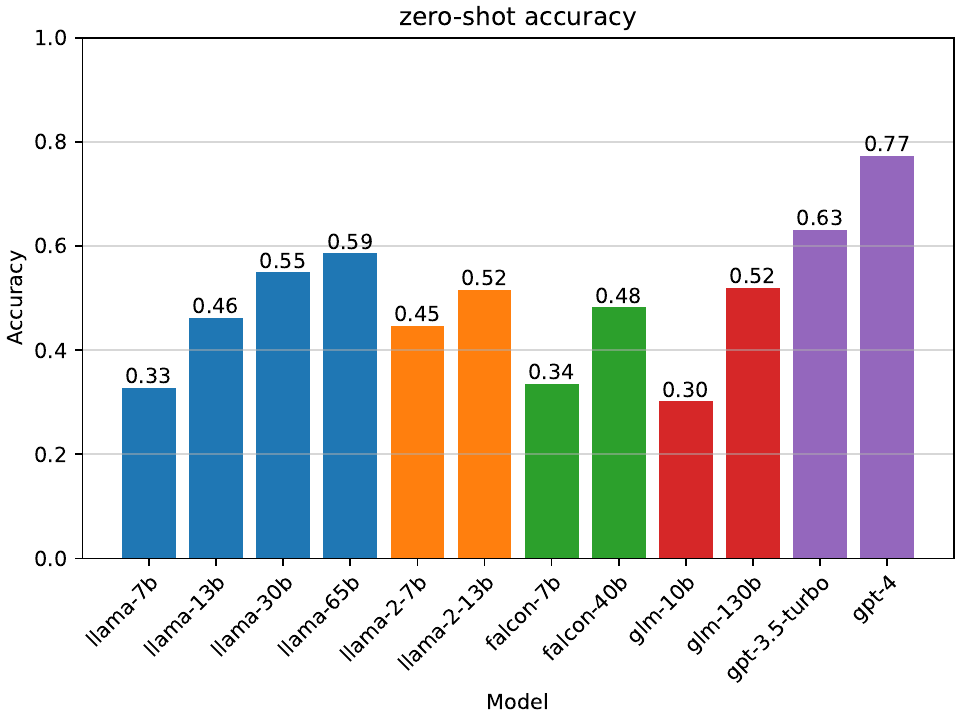} %
  \caption{Zero-shot Accuracy of Models with Different Scales within the Same Category}
  \label{fig:image3}
\end{figure}

As depicted in Figure \ref{fig:image3}, a comparison of accuracy across different parameter scales of the same model reveals that larger size LLMs perform better. When comparing different models, we observe that LLaMA 2 70B is the best-performed open model that is competitive with GPT-4. Furthermore, accuracy of LLaMA 65B is greater than GLM 130B, which has more than twice the parameter count. This could be attributed to the scarcity of NetOps-related English language data in the training corpus of domestic models.

\subsection{Impact of Zero-shot/Few-shot Prompts on Accuracy}
\begin{figure}[h]
  \centering
  \includegraphics[width=\textwidth]{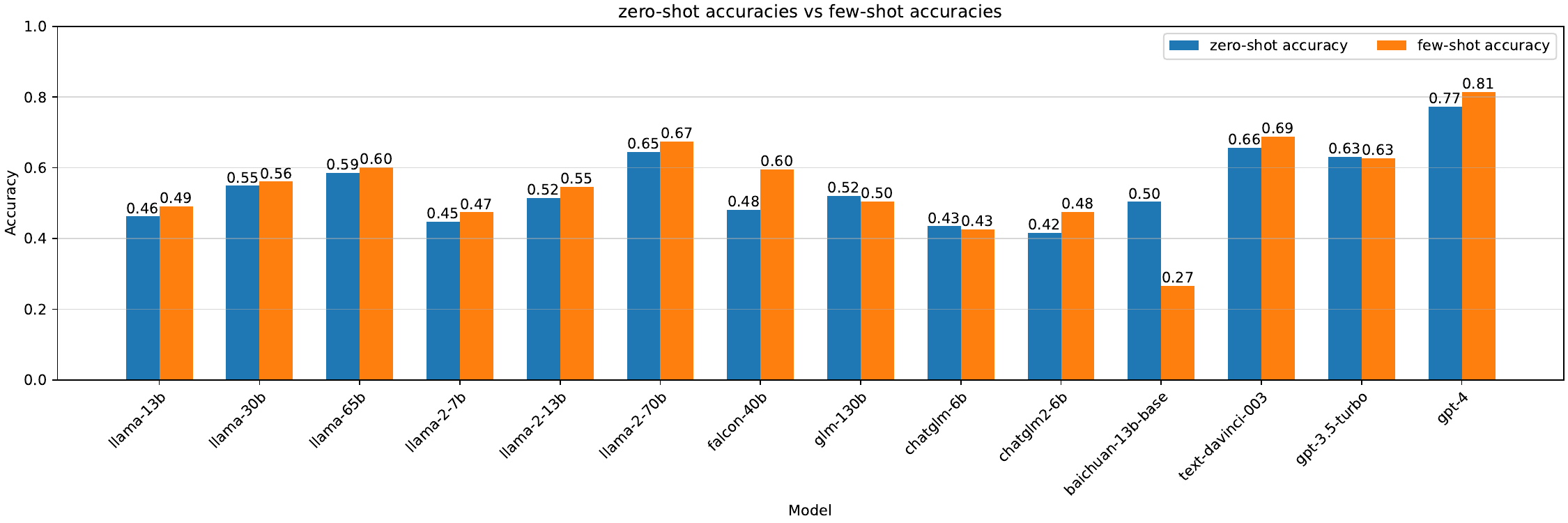} %
  \caption{Comparison of Zero-shot and Few-shot Accuracy}
  \label{fig:image4}
\end{figure}

With LLMs getting ever larger, prompt engineering is a powerful tuning method for a specific task. We compared the two common prompts settings, zero-shot and few-shot prompt.
The result is shown in Figure \ref{fig:image4}, we only display the accuracy of models which obtain at least 40\% accuracy in zero-shot setting. It is found that the effect of few-shot prompts on model accuracy varies depending on the model type. For LLaMA, LLaMA-2, GLM-130B, ChatGLM-6B, and GPT-3.5, few-shot prompts had a mild impact (<3\%).  Falcon-40B and ChatGLM2-6B exhibited substantial improvement, with an improvement of more than 6\%. Falcon-40B also achieved comparable performance to that of LLaMA-65B. However, for GLM-130B, few-shot prompts had a negative impact. It possibly dues to its limitation prompt understanding in English. The examples in the prompt seem to mislead the model instead.

\subsection{Impact of Instruction Tuning on Accuracy}

\begin{figure}[ht]
  \centering
  \includegraphics[width=0.7\textwidth]{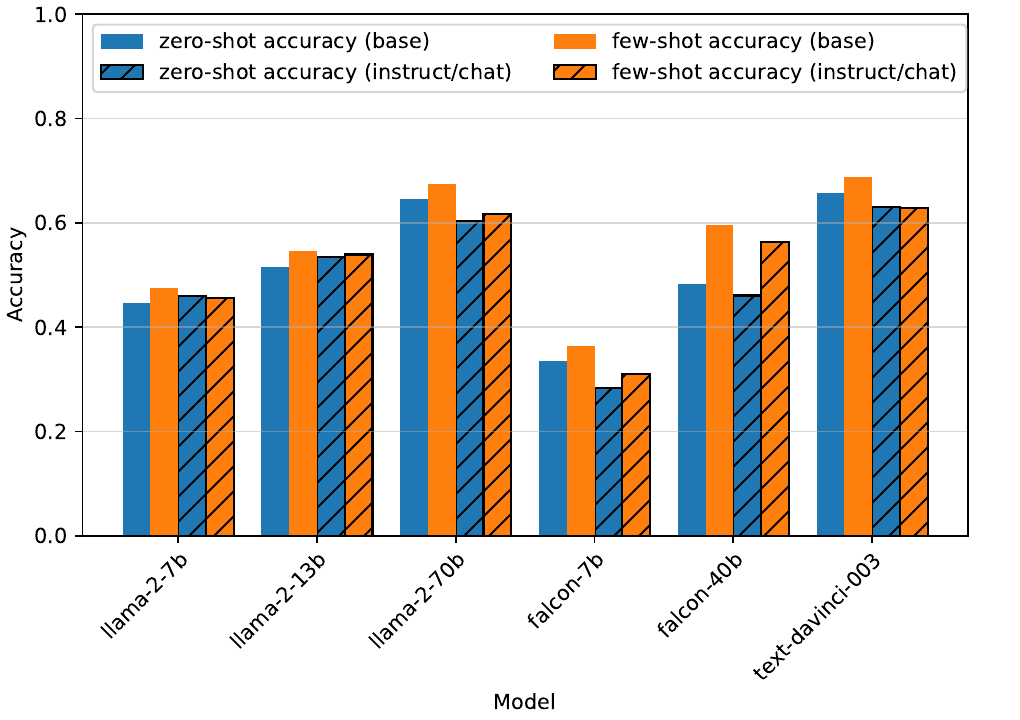} %
  \caption{Impact of Instruction Tuning with general domain corpus}
  \label{fig:image5}
\end{figure}

Instruction Tuning (IT) or Supervised Fine-Tuning(SFT) is a common technique used to further align models to tasks after pretraining. We aimed to assess whether IT is beneficial for models in the NetOps domain. As shown in Figure \ref{fig:image5}, we compared the performance of original versions and the IT versions for both LLaMA-2 and Falcon models. Interestingly, for both LLaMA-2 and Falcon, the models after IT did not perform as well as the original models. This suggests that IT might not necessarily improve model performance. This outcome could be attributed to the fact that the IT data used for LLaMA-2-Chat and Falcon-instruct may have substantial differences from the domain knowledge of NetOps. Therefore, it is necessary to employ domain-specific data for IT to enhance model effectiveness in NetOps field.

\subsection{Effect of Chain-of-Thought}

\begin{figure}
    \centering
    \includegraphics[width=0.6\textwidth]{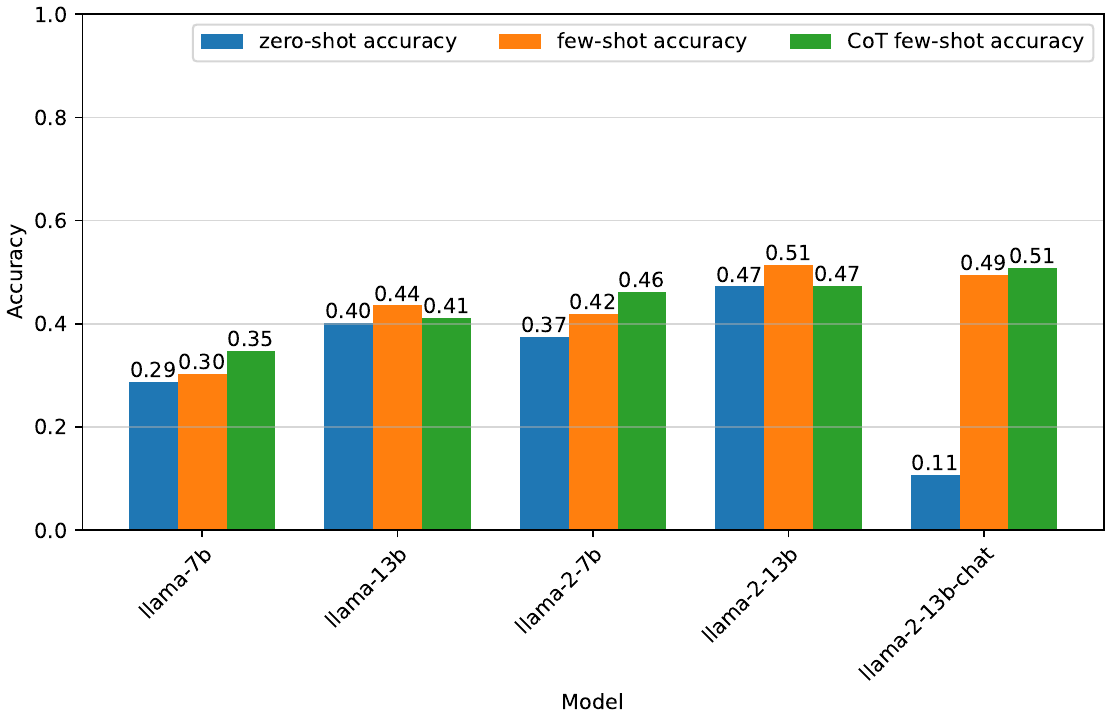}
    \caption{Results on Chain-of-Thought prompts}
    \label{fig:cot-result}
\end{figure}

As CoT substantially increase the length of both the prompt and the output, which consumes much more computation resources than non-CoT settings, we only evaluate the CoT prompt in selected models. As shown in Figure~\ref{fig:cot-result}, LLaMA-7B and LLaMA-2-7B obtain 4-5\% increase in few-shot accuracy when CoT is applied, while other models do not have positive results. This suggests that few-shot CoT
might be more effective on smaller models. 

\subsection{Effect of Retrieval-Augmented Generation}

\begin{figure}
    \centering
    \includegraphics[width=0.6\textwidth]{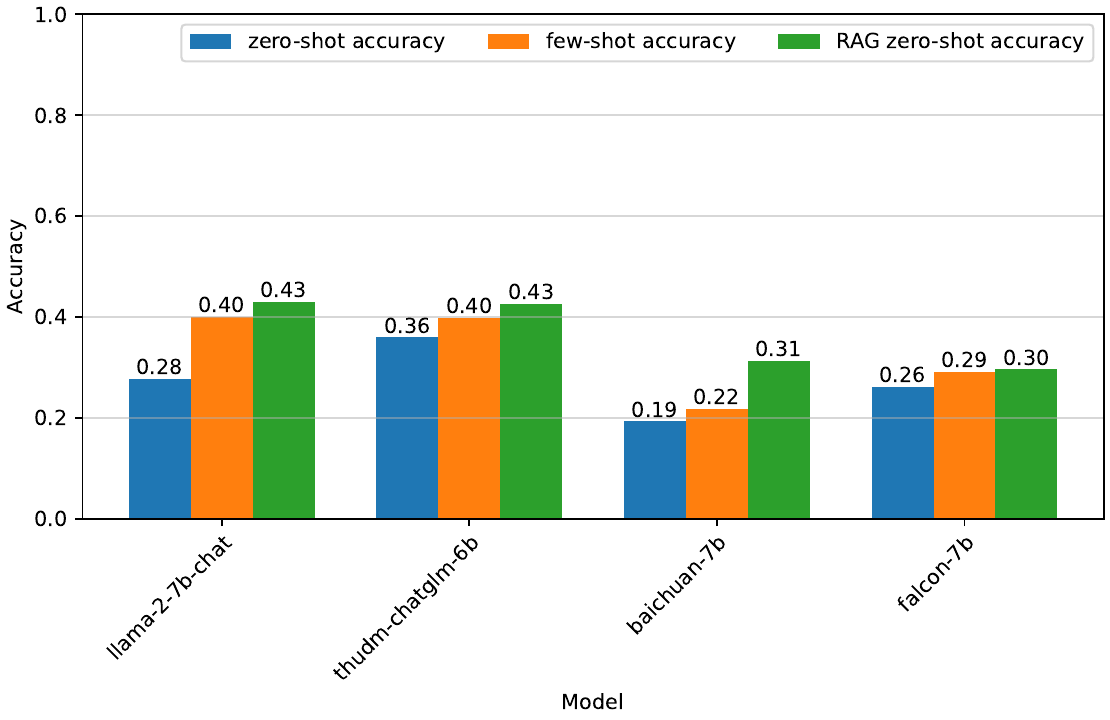}
    \caption{Results on Retrieval-Augmented Generation}
    \label{fig:rag-result}
\end{figure}

Like CoT prompt, RAG prompt incurs high cost as well. 
Thus, as shown in Figure~\ref{fig:rag-result}, we only evaluate 4 models on RAG prompts. We find that for all the evaluated models, RAG consistently improves the zero-shot performance and even achieves better results than few-shot prompts without RAG. This result is consistent with our expectation that RAG enriches the context information with external relevant knowledge and facilitates the model inference.

\subsection{Impact of Question Language on Accuracy}

\begin{figure}
    \centering
    \includegraphics[width=\textwidth]{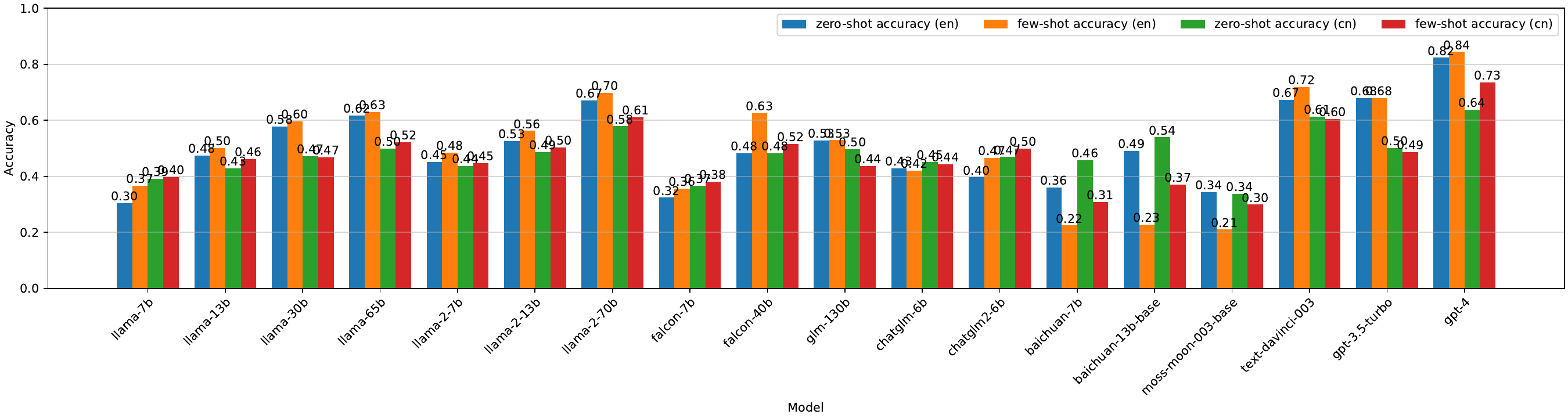}
    \caption{Accuracy on questions of different languages}
    \label{fig:question_lang_result}
\end{figure}

The distribution of languages in the pre-training data may affect the performance of the pre-trained LLM on multi-lingual tasks. As shown in Figure~\ref{fig:question_lang_result}, we compute the model accuracy on English questions and Chinese questions separately to measure the multi-lingual NetOps capability. Except GLM-130B, models pre-trained with Chinese/English bilingual corpus like Baichuan, Moss, and ChatGLM 2 achieve higher accuracy on Chinese questions than on English questions. For instance, Moss-moon-sft under zero-shot prompt and Baichuan-13B-base under few-shot prompt achieve 14 to 17 percents higher accuracy on Chinese questions than the accuracy on English questions. For other models, the accuracy on Chinese questions is lower than the accuracy on English questions. For GPT-4, the zero-shot accuracy on Chinese questions is 18 percents lower than that on English questions, which suggests that there is still substantial space for improvement on the Chinese NetOps capabilities of LLMs.

\subsection{Evaluation on Non-multi-choice Questions}

\begin{figure}
    \centering
    \includegraphics[width=\linewidth]{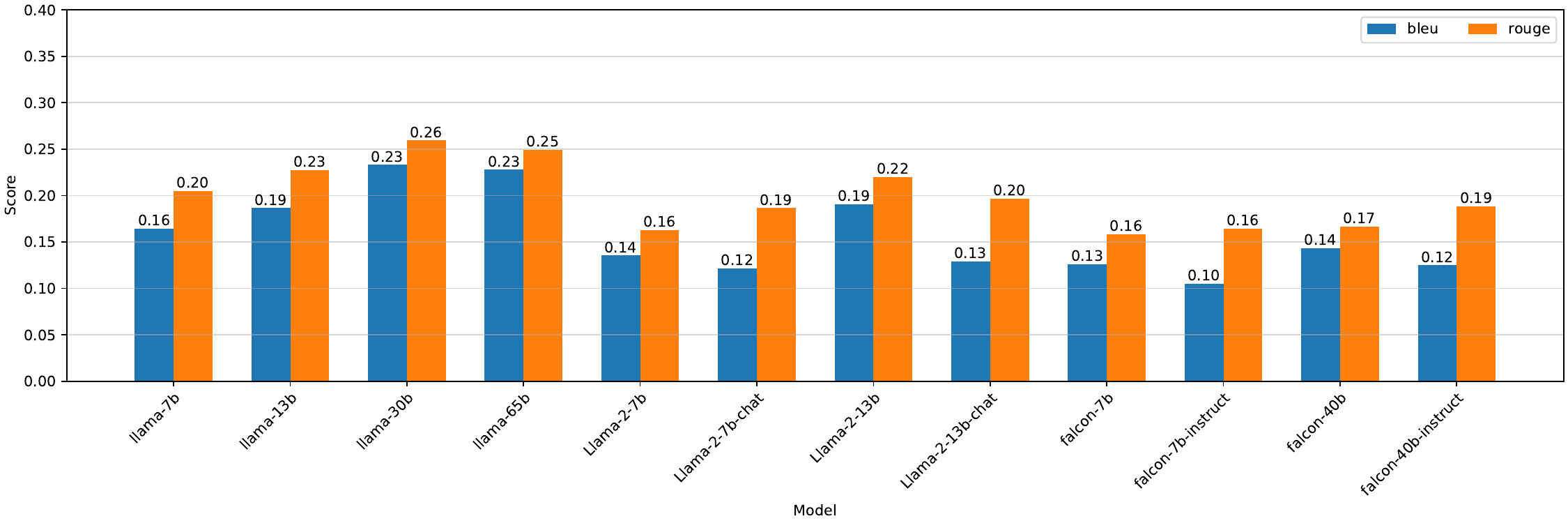}
    \caption{Results on Non-multi-choice Questions}
    \label{fig:qa-results}
\end{figure}

As shown in Figure~\ref{fig:qa-results}, we evaluate 12 models on non-multi-choice questions. We adopt zero-shot prompt in this evaluation. Overall, all the evaluated models do not achieve good performance as their BLEU/ROUGE scores are all below $0.3$, which shows that this evaluation set is hard. We find that LLaMA models have better performance than the other evaluated models, including LLaMA 2. It is likely that LLaMA 2 is not as good as LLaMA in closed-book question-answering. For LLaMA-2-7B and Falcon, comparing their base model and the chat/instruct model, we see that the chat/instruct model achieves lower precision but higher recall.

\section{Conclusion and Future Work}\label{sec:conclusion}

In this report, we present a comprehensive multi-lingual LLM evaluation set for NetOps, named NetEval, and systematically evaluate the NetOps capabilities of 26 widely used LLMs . The results show that only GPT-4 is capable of reaching human-level performance, which reveals both the huge potential and the space for improvement of LLMs in NetOps domain.

In the future, we will keep improving the quantity, quality and diversity of the questions in NetEval, involve more LLMs in the evaluation, and share the results and insights with the community. Furthermore, as NetOps is a subject emphasizing on practice, we are considering adding to NetEval some practical NetOps tasks, which are executed and evaluated in simulated networking environments.

\bibliographystyle{unsrt}  
\bibliography{references}

\end{document}